\documentclass[format=sigconf, authorversion=true]{acmart}

\usepackage[disable]{todonotes}
\usepackage{siunitx} 
\usepackage{mathtools} 
\usepackage{amsmath}
\usepackage{amsfonts}
\usepackage{amssymb}

\usepackage{enumerate}
\usepackage{bm}
\newcommand{\trsp}{{\!\scriptscriptstyle\top}}

\newcommand{\reals}{{\mathbb{R}}} 

\newcommand{\refig}[1]{{Fig. \ref{#1}}}
\newcommand{\refsect}[1]{{\S \ref{#1}}}

\renewcommand{\vec}{\bm} 
\newcommand\given[1][]{\:#1\vert\:}

\copyrightyear{2017}
\acmYear{2017}
\setcopyright{acmcopyright}
\acmConference{MOCO '17}{June 28-30, 2017}{London, United Kingdom}\acmPrice{15.00}\acmDOI{10.1145/3077981.3078049}
\acmISBN{978-1-4503-5209-3/17/06}

\begin{document}

\title[Calligraphic Stylisation Learning based on Movement and RNNs.]{Calligraphic Stylisation Learning with a Physiologically Plausible Model of Movement and Recurrent Neural Networks}

\author{Daniel Berio}
\affiliation{%
  \institution{Goldsmiths, University of London}
}
\email{d.berio@gold.ac.uk}

\author{Memo Akten}
\affiliation{%
  \institution{Goldsmiths, University of London}
}
\email{m.akten@gold.ac.uk}

\author{Frederic Fol Leymarie}
\affiliation{%
  \institution{Goldsmiths, University of London}
}
\email{ffl@gold.ac.uk}

\author{Mick Grierson}
\affiliation{%
  \institution{Goldsmiths, University of London}
}
\email{m.grierson@gold.ac.uk}

\author{R\'ejean Plamondon}
\affiliation{%
  \institution{Ecole Polytechnique de Montr\'eal}
}
\email{rejean.plamondon@polymtl.ca}


\begin{abstract}
We propose a computational framework to \textit{learn} stylisation patterns from example drawings or writings, and then generate new trajectories that possess similar stylistic qualities. We particularly focus on the generation and stylisation of trajectories that are similar to the ones that can be seen in calligraphy and graffiti art. Our system is able to extract and learn dynamic and visual qualities from a small number of user defined examples which can be recorded with a digitiser device, such as a tablet, mouse or motion capture sensors. Our system is then able to transform new user drawn traces to be kinematically and stylistically similar to the training examples. We implement the system using a Recurrent Mixture Density Network (RMDN) combined with a representation given by the parameters of the Sigma Lognormal model, a physiologically plausible model of movement that has been shown to closely reproduce the velocity and trace of human handwriting gestures.
\end{abstract}

%
%

\ccsdesc[500]{Computing methodologies~Neural networks}
\ccsdesc[500]{Computing methodologies~Procedural animation}
\ccsdesc[300]{Human-centered computing~Gestural input}


\keywords{Human hand-writing movement modeling; procedural calligraphic and graffiti production; graffiti tags; curve and path dynamics; Recurrent Neural Network, RNN; Long Short-Term Memory architecture, LSTM; Mixture Density Network, MDN; Recurrent Mixture Density Network, RMDN; graphonomics; one-shot learning.}

\maketitle

\section{Introduction}

When writing or drawing, the pen or brush leaves a static mark which is a record of the movement underlying its production. Each such static trace inherits qualities directly due to the generating movement, and is further unique to each individual. There is cumulating cognitive and neuroscientific evidence to suggest that, as observers, humans implicitly (at a neural level) reconstruct a movement when observing such traces on a canvas \cite{pignocchi2010,Umilta2012}. Furthermore, while a given letter may be written with different topological structures --- consider for example a lower case "a" written as a cursive "$\alpha$" --- and different ordering of generative gestures, the same topology and gesture ordering may themselves be executed with a possibly infinite number of variations. Such variations are principally determined by the movement dynamics used in the production of a letter. In our study we will consider a particular model of movement when writing as a powerful descriptor of various stylisations when applied to a common topological structure of a letter form.

\begin{figure}[ht] 
	\vspace{-2mm}
	\centering
	\includegraphics[width=0.5\textwidth]{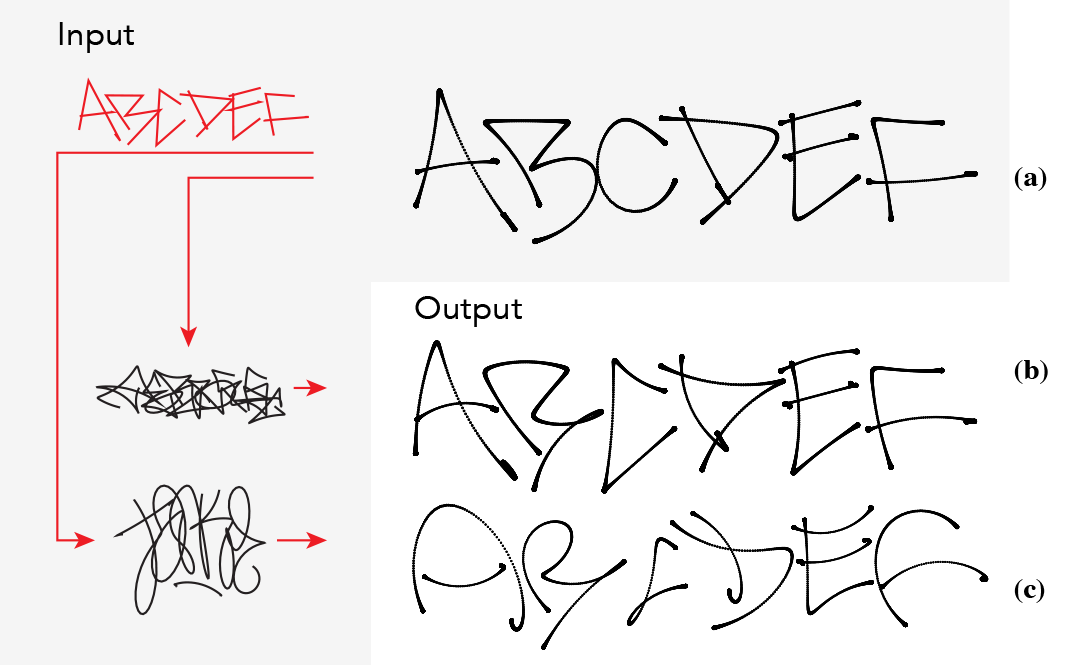} 
	\caption{Stylisation of a user drawn trace and one-shot learning. (a) An input trace drawn by a user (left) and the extracted virtual targets (right, in red). (b,c) We train two DPP models, each on a single example (right). Each model predicts new dynamic parameters for the user drawn trace (a) to generate the trajectory on the left.}
	\label{fig:style_transfer}
\end{figure}

We base our work on the Sigma Lognormal model which is recognised as a physiologically plausible representation of the kinematics of hand movements \cite{Plamondon2014}. This model facilitates the implementation of a bi-level representation of a written trace made of a \textit{structural} and a \textit{dynamic} component. The \textit{structural} component defines the overall spatial layout of the handwriting trace through a coarse ordered sequence of targets. Such targets define the aiming position for a sequence of consecutive goal directed sub-movements (\textit{aka} movement primitives or strokes). The \textit{dynamic} component determines the velocity, acceleration, timing and curvature of each sub-movement, and produces a smooth trajectory that follows the spatial layout of the targets. By exploiting this bi-level representation, we train a type of Recurrent Neural Network (RNN) that learns correlations between the structure of a trace and the corresponding parameters that control its fine dynamic evolution (\refig{fig:style_transfer}).

To learn these correlations, we use \textit{Recurrent Mixture Density Networks (RMDNs)} with \textit{Long Short-Term Memory (LSTM)} architecture. Recent developments have shown that such types of RNNs are capable of modelling complex sequential (or time-ordered) data such as text \cite{Sutskever2013}, images \cite{Gregor2015}, dance \cite{friis2016} and handwriting \cite{Graves2013}. While most existing approaches aim to minimise the use of "hand-crafted features", we hypothesise that for our task it is beneficial to formulate a mid-level mapping that exploits our knowledge of the specific problem domain. Our rationale is that human handwriting (and related artistic processes) results from the orchestration of a large number of motor and neural subsystems, and that movement is arguably planned using some form of higher level mapping, possibly in the form of movement primitives which are then combined in a syntactic manner similar to language \cite{Flash2005}. As a result, for this particular study, we seek a mapping that abstracts the complex task of trajectory formation from the neural network, which is then left with focusing on the higher level task of movement planning. 
	
In this paper, we demonstrate our approach on digitised traces, which are recorded by a user with a pen-tablet or downloaded from a large online graffiti motion database (GML, \cite{GraffitiAnalysis}). We first transform the input into an intermediate representation given by the parameters of the Sigma Lognormal model \cite{Plamondon2014}. Doing so requires a preprocessing step which results in a representation that is more concise (\textit{i.e.} with low cardinality) and meaningful, such that every representative locus is now a high level segment of the trajectory when compared to an initial sequence of positions. Such a representation is also \textit{resolution independent}, and can easily be manipulated prior and after training. For example, it is possible to introduce random perturbations at the model parameter level, which results in new samples that mimic the variability one might observe when an artist draws the same form multiple times. We exploit this capability to augment our training data and learn from very small amounts of input data-sets --- as small as a single training example, as we will demonstrate (aka one-shot learning).
	
In the following, after summarising the background (\refsect{sect:background}), we briefly describe the Sigma Lognormal model, present the data preprocessing step and the implemented RMDN model (\refsect{sect:method}); then we propose various applications of the system, including learning handwriting representations from small datasets and mixing styles (\refsect{sect:results}).

\section{Background}
\label{sect:background}

Our study is grounded on a number of notions and principles that have been observed in the general study of human movement as well as in \textit{graphonomics}, the handwriting synthesis and analysis scientific field  \cite{kao1986graphonomics}. The speed profile of aiming movements is  typically characterised by a ``bell shape'' that is variably skewed depending on the speed 
\cite{Nagasaki1989}. Complex movements can be described by the superimposition of a discrete number of ``ballistic'' units of motion, which in turn can each be represented by the classic bell shaped velocity profile and are often referred to as \textit{strokes}. A number of methods synthesise handwriting through the temporal superimposition of strokes, the velocity profile of which is modelled with a variety of functions including sinusoidals
\cite{Rosenbaum1995}, Beta functions 
\cite{bezine2004beta}, and lognormals \cite{plamondon1995kinematic}. 

In our work we rely on a family of models known as the \textit{Kinematic Theory of Rapid Human Movements}, mainly developed by R. Plamondon \textit{et al.} in an extensive body of work since the 90's  \cite{plamondon1995kinematic,Plamondon2014}. They show that if we consider that a movement is the result of the parallel and hierarchical interaction of a large number of coupled linear systems, the impulse response of such a system to a centrally generated command asymptotically converges to a lognormal function. This assumption is attractive from a modelling perspective because it abstracts the high complexity of the neuromuscular system in charge of generating movements with a relatively simple mathematical model, which further provides state of the art reconstruction of human velocity data \cite{plamondon1995kinematic,Rohrer2006}.

A number of methods have used neural inspired approaches for the task of handwriting trajectory formation.
With a preprocessing step similar to ours, Ltaief et al.~\cite{Ltaief2012} use a neural network to learn the mapping between the parameters of a handwriting model \cite{bezine2004beta} and the corresponding trajectory. The network is then used instead of the model as a trajectory generator. Nair and Hinton \cite{Nair2005} use a sequence of neural networks to learn the motion of two orthogonal mass spring systems from images of handwritten digits. The authors also propose that a motor representation can be exploited to generate realistically augmented training sets by introducing noise at the motor level. Plamondon and Privitera \cite{Plamondon1996}
use a Self Organising Map (SOM) to learn a sparse sequence of ballistic targets, which is then used in conjunction with a Kinematic Theory model to generate trajectories. More recently, Graves \cite{Graves2013} has demonstrated that the LSTM RNN combined with MDNs and trained on a large online dataset, are able to synthesise convincing handwriting sequences. His method provides the basis for a number of extensions \cite{ha2016hypernetworks,Zhang2016}. In our contribution, we design an RNN architecture inspired by the one studied by Graves together with a motor representation similar to the one of Plamondon and Privitera \cite{Plamondon1996}, and with an original goal of \textit{trajectory stylisation}.
While Graves works with a dense sequence of points sampled at fixed frequency, our method operates on a sparse and concise representation, which provides more flexibility and allows for meaningful manipulations that would be difficult to achieve otherwise. We demonstrate and discuss these in the remainder of the paper.

In the computer graphics domain, a number of data driven methods have been proposed for the task of vector path stylisation (e.g.
\cite{hertzmann2002curve,lang2015markov}). These method consider stylisation as a bi-level transfer problem: given two sketches, generate a third one with the structure or "content" of the first but with a style similar to the second. Our approach is also organised to function via a bi-level mapping, but focused on the motor level, such that we represent (i) "content" with a coarse structural description of a drawn trace, and (ii) "style" with different types of movements that follow the same structure. In the work reported here, we approach the problem with a data driven method, and learn such a bi-level mapping from user provided examples. In a complementary line of work \cite{BerioMPCMOCO2017,BerioGI2017}, we approach a similar problem from a procedural perspective, and use optimal control to define different stylisations of a given geometric structure.

\section{Method}
\label{sect:method}
\begin{figure}[ht] 
	\vspace{-3mm}
	\centering
	\includegraphics[width=0.5\textwidth]{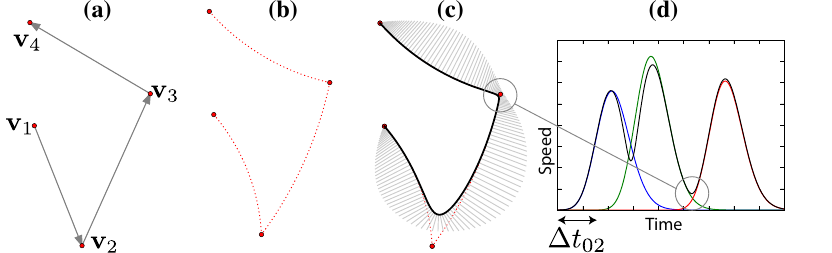} 
	\caption{A sequence of virtual targets and the corresponding $\Sigma \Lambda$ trajectory. \textbf{(a)} Virtual targets and corresponding stroke aiming directions. \textbf{(b)} Corresponding circular arcs. \textbf{(c)} Possible trajectory generated over the given sequence of virtual targets. While the generated trajectory appears visually similar to an interpolating spline, it further describes a smooth and physiologically plausible \textit{velocity} profile \textbf{(d)}.}
	\label{fig:action_plan}
\vspace{-3mm}
\end{figure}

Our intermediate representation relies on the Sigma Lognormal ($\Sigma \Lambda$) model \shortcite{plamondon1995kinematic,Plamondon2014}, which generates trajectories that closely resemble the geometry \shortcite{Plamondon2014} and kinematics \shortcite{Plamondon1993, Rohrer2006} of those made by an experienced human writer. The evolution of a pen-tip trajectory is then described with the spatio-temporal superimposition of a discrete number of ballistic strokes, where each consecutive stroke is aimed at an imaginary position denoted as a \textit{virtual target}. We define our bi-level representation with (i) the structural component given by a \textit{sequence of virtual targets}, and (ii) the dynamic component given by a subset of the remaining $\Sigma \Lambda$ parameters which define the fine evolution of the trajectory and which we refer to as \textit{dynamic parameters} (\refig{fig:action_plan}). 

The input to the system is given by a discrete and temporally ordered sequences of planar coordinates which are recorded with a digitizer device, such as a tablet or mouse. As a preliminary step, we preprocess this input data and reconstruct it in the form of $\Sigma \Lambda$ model parameters (\refsect{sect:preprocessslm}). 

We then use this representation to train an RMDN model that learns to predicts the dynamic parameters corresponding to a given sequence of virtual targets (\textit{Dynamic Parameter Prediction} or DPP, \refsect{sect:DPP}). In order to train on small datasets, we augment the preprocessed data by introducing artificial variability at the $\Sigma \Lambda$ parameter level.
In the following sections we describe the steps that constitute our system and then demonstrate its use case with a number of examples (\refsect{sect:results}).

\subsection{Sigma lognormal model}
\label{sect:slm}

The $\Sigma \Lambda$ model explicitly describes the velocity profile of a stroke with a $3$ parameter lognormal function:
   		\begin{equation}
   		\label{eq:lognormal}
   		\Lambda_i (t) = -\frac{1}{\sigma_i \sqrt{2\pi}(t-{t}_{0i})}
		\mathrm{exp}\left(\frac{ {(ln(t-{t}_{0i}) - \mu_i)}^{2} }{2 {\sigma_i}^{2}} \right)  ,  
	\end{equation}
where ${t}_{0i}$ defines the activation time of a stroke and the parameters $\mu_i$ and $\sigma_i$ determine the shape of the lognormal function. A detailed discussion of the biological interpretation of these parameters is presented by Plamondon \cite{plamondon1995kinematic}.
\textit{Assuming} that curved handwriting movements result from rotating the wrist, strokes are described by circular arcs:
\begin{equation}
	\label{eq:sigma_angle}
	{\phi}_{i}(t) = \theta_i + \theta_i
	\left[ 1 + \mathrm{erf}\left( \frac{\mathrm{ln}(t - {{t}_{0}}_{i}) - {\mu}_{i} } { {\sigma}_{i} \sqrt{2} } \right) \right],
\end{equation}%
where $\theta_i$ is the central angle of the arc for the $i$th stroke.
A complete trajectory with $m$ strokes is defined with an initial position $\bm{v}_0$ followed by a \textit{sequence of virtual targets} $\left\{\bm{v}_i\right\}_{i=1}^{i=m}$, where a trajectory with $m$ virtual targets will be characterised by $m-1$ circular arc strokes. The trace can be then integrated with:
\begin{align}
	\label{eq:sigma_lognormal}
	\bm{\xi}(t) &=\bm{v}_1 +\int _{ 0 }^{ t }{ d\tau
	\Lambda_i(\tau)
	\sum _{ i=1 }^{ m-1 }{ 
		\bm{\Phi}_i(\tau)
		\left( \bm{v}_{i+1} - \bm{v}_i \right)
		}
		}, \\
\text{where}\quad \bm{\Phi}_i(t) &= \begin{bmatrix} {h(\theta_i)} \mathrm{cos} \phi_i(t) & 
						-{h(\theta_i)} \mathrm{sin} \phi_i(t) \\
						{h(\theta_i)} \mathrm{sin} \phi_i(t) & 
						-{h(\theta_i)} \mathrm{cos} \phi_i(t) 
		\end{bmatrix},\\
		\text{and} \quad {h(\theta_i)} &= \begin{cases}
	 \frac{2 \theta_i}{  2 \mathrm{sin} \theta_i  } &\text{if $\left| \mathrm{sin} \theta_i  \right| > 0$},\\
	1 &\text{otherwise},
	\end{cases}
\end{align}
which scales the extent of the stroke based on the ratio between the perimeter and the chord length of a circular arc.

\noindent\textbf{Explicit asymmetry parameterisation.} In order to facilitate the precise specification of timing and profile of each stroke, we explicitly define the structure of each lognormal profile through its duration $T_i$, a skewness parameter $A_{c_i}$ and a time offset $\Delta t_i$ with respect to the onset of the preceding stroke. The corresponding $\Sigma \Lambda$ parameters $\left\{ t_{0i}, \mu_i, \sigma_i \right\}$ are then obtained as:
\begin{align}
	\sigma_i = \sqrt{- \log ( 1 - A_{c_i} ) } , \qquad \mu_i &= 3 \sigma_i - \log( \frac{ -1 + e^{6\sigma_i} }{T_i} ) ,
\end{align}
\begin{equation}
t_{0i} = t_{1i} - e^{\mu - 3 \sigma} , \qquad
t_{1i} = t_{1(i-1)} + d_{i-1} \Delta t_i  , \qquad
t_{1(0)} = 0 ,
\end{equation}
where $t_{1i}$ is the onset time of the lognormal stroke profile.

\noindent\textbf{Dynamic parameters.}
In our method, we use the pairs $(\Delta t_i, \Theta_i)$ as the \textit{dynamic parameters} that are mapped to a corresponding sequence of virtual targets. We keep the remaining parameters $A_{c_i}$ and $T_i$ (or accordingly $\mu_i, \sigma_i$) fixed to predefined values (experimentally set to $0.1$ and $0.3$). While this is a simplifying assumption, these parameters mostly determine temporal properties of the writing motion, and thus do not produce a strong effect on the shape of the resulting trajectory. 
\subsection{Preprocessing}
\label{sect:preprocessslm}
A number of methods have been developed by Plamondon \textit{et al.} in order to reconstruct $\Sigma \Lambda$-model parameters from digitised pen input data \cite{OReilly2008,Plamondon2014,Fischer2014}. These methods provide the ideal reconstruction of model parameters, given a high resolution digitised pen trace. While such methods are superior for handwriting analysis and biometric purposes, we opt for a less precise method \cite{berio2015cae} that is less sensitive to sampling quality and is aimed at generating virtual target sequences that remain perceptually similar to the original trace. We purposely choose to \textit{ignore the original dynamics} of the input, and base the method on a geometric trace only.  This is done in order to work with training sequences that are independent of sampling rate.

Our method operates on an uniformly sampled input trace $\bm{p}[k]$, where the sampling distance depends on the maximum extent of the trace's bounding box. As an initial step (i) the trace is segmented in correspondence with perceptually salient \textit{input key points}: i.e. loci of curvature extrema modulated by neighbouring contour segments
\cite{berio2015cae}. The key points define the number of strokes in the $\Sigma \Lambda$ trajectory and provide an initial estimate of virtual target positions. We then (ii) fit a circular arc to each segment in order to estimate the $\theta_i$ parameters and (iii) estimate the $\Delta t_{0i}$ parameters by analysing the trace curvature in the region of each key point. Finally, (iv) we iteratively adjust the virtual target positions to minimise the error between the original trajectory and the one generated by the corresponding $\Sigma \Lambda$ parameters. In the following paragraphs we describe the steps in more detail. 

\noindent\textbf{(i) Estimating input key-points.}
To compute salient points along the input trace we rely on a method described by Fedman and Singh \cite{feldman2005information} and supported experimentally by De Winter \textit{et al.} \cite{dewinter2008perceptual}. We then find local maxima in the signal:
\begin{equation}
\label{eq:turning_angle}
s[k] = 1 - \mathrm{cos}(\theta[k]) ,
\end{equation}
where $\theta[k]$ is the turning angle of the $k$th sample of the input, convolved with a Hanning window of size $40$ to smooth noise in the input. The first and last sample indices of $s[k]$ plus its local maxima results in $m$ key-point indices $\left\{ \hat{z}_i \right\}$ with coordinates $\left\{ \bm{p} \left[ \hat{z}_i \right] \right\}$.

\begin{figure}[h]
\centering
\includegraphics[width=0.5\textwidth]{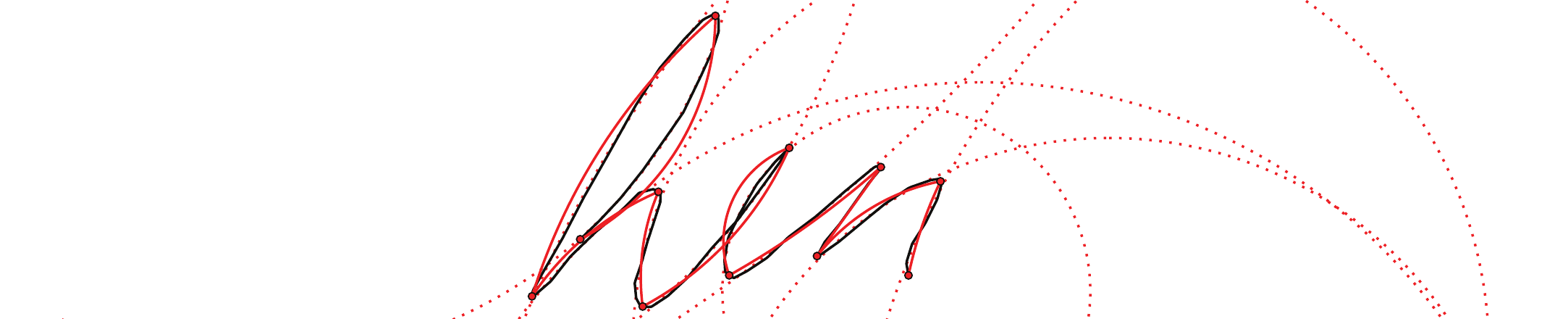} 
\caption{Fitting arcs (and associated circles as dotted lines).}
\label{fig:arc_fit}
\end{figure}

\noindent\textbf{(ii) Estimating stroke curvature parameters.} For each section of the input trace defined between two consecutive key-points, we estimate the corresponding stroke curvature parameter $\theta_i$ by first computing a least square fit of a circle to the trace segment (\refig{fig:arc_fit}). We then compute the internal angle of the arc supported between the two key-points, which is equal to $2 \theta_i$.

\begin{figure}[h]
\centering
\includegraphics[width=0.5\textwidth]{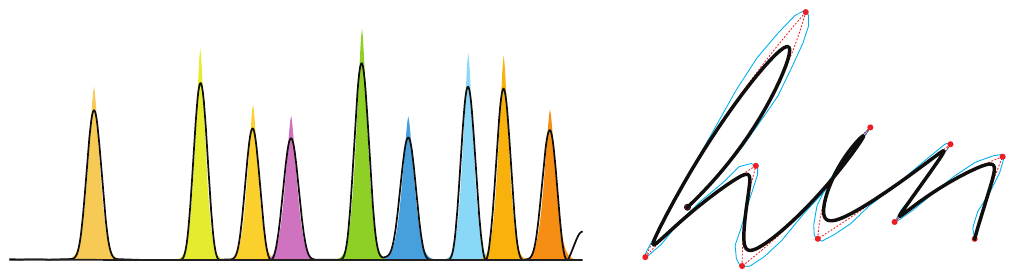} 
\caption{Sharpness estimation. Left, the GMM components (coloured) estimated from the turning angle signal (black) computed with Eq. \eqref{eq:turning_angle}. Right, the $\Sigma \Lambda$ trajectory generated before the final iterative adjustment step. Note that at this stage the virtual target positions correspond with the estimated input key-points.}
\label{fig:sharpness}
\end{figure}

\noindent\textbf{(iii) Estimating stroke time-overlap parameters.}
This step is based on the observation that smaller values of $\Delta t_{0i}$, \textit{i.e.} a greater time overlap between strokes, result in smoother trajectories. On the contrary, a sufficiently large value of $\Delta t_{0i}$ will result in a sharp corner in proximity of the corresponding virtual target. We exploit this property to compute an estimate of the $\Delta t_{0i}$ parameters by examining the sharpness of the input trace in the neighbourhood of each key-point. 
To do so we consider the previously computed signal $s[k]$ (\refeq{eq:turning_angle}), in which we can observe that sharp corners correspond with sharp signal peaks, while smoother corners correspond with smooth peaks having a larger spread. By treating the signal as a probability density function, we can then employ a variant of Expectation Maximisation (EM) \cite{Dempster77} to measure the shape of each peak with a mixture of parametric distributions (\refig{fig:sharpness}) --- the reader is referred to recent work by Berio \textit{et al.} \cite{Berio16IROS} for additional details on the EM variant. We then examine the shape of each mixture component in order to get an estimate of the corresponding sharpness along the input trace. We further generate an estimate of sharpness $\lambda_i$ (bounded in the $[0,1]$ range) for each key point using a logarithmic function of the mixture parameters and weights. The corresponding $\Delta t_{0i}$ parameters are then given by:
\begin{equation}
		{\Delta t}_{i} = {\Delta t}_{min} + ({\Delta t}_{max}-{\Delta t}_{min}) \lambda_i \ \ ,	
		\label{eq:dtj}
	\end{equation}
where ${\Delta t}_{min}$ and ${\Delta t}_{max}$ are user specified parameters that determine the range of the $\Delta t_{0i}$ estimates.
For the sharpness estimation function we currently use a mixture of Gaussians and empirically defined thresholds for the estimation $\lambda_i$. While this gives satisfactory results for our use-case, further studies are needed to limit the number of free parameters for this step. In future work we plan to test different types of distributions, such as generalised Gaussians, to learn the mapping between sharpness and mixture component parameters from synthetically samples generated with the $\Sigma \Lambda$ model, for which $\Delta t_{0i}$, and consequently $\lambda_i$, are known.
  
\begin{figure} 
\centering
\includegraphics[width=0.5\textwidth]{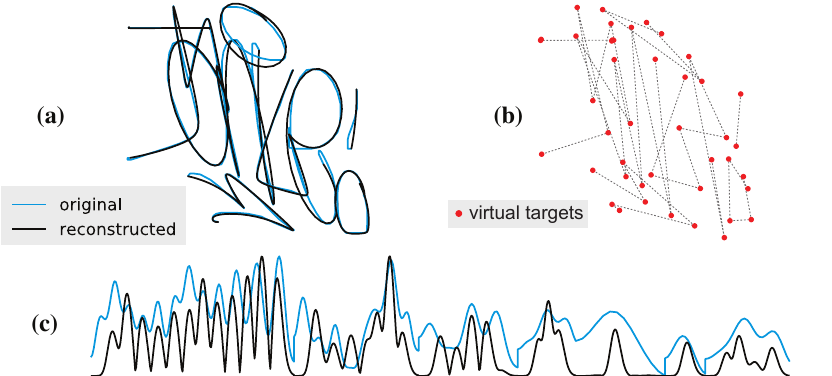} 
\caption{$\Sigma \Lambda$ parameter reconstruction. (a) Original (black) and reconstructed (turquoise) trajectories. (b) Reconstructed virtual targets. (c) Aligned and scaled speed profiles of the original and reconstructed trajectories.}
\label{fig:param-recons}
\end{figure}

\noindent\textbf{(iv) Iteratively estimating virtual target positions.} 
The loci of the input key-points provide an initial estimate for a sequence of virtual targets, where each virtual target position is given by ${\bm{v}}_{i}=\bm{p}[\hat{z}_i]$. Due to the trajectory-smoothing effect produced by the time overlaps, the initial estimate will result in a generated trajectory that is likely to have a reduced scale with respect to the input we wish to reconstruct \cite{varga2005template}. In order to produce a more accurate reconstruction we compute an estimate of $m$ \textit{output} key-points $\left\{ \bm{\xi} \left( z_i \right) \right\}$ in the generated trajectory, where $z_2, ..., z_m$ are the time occurrences at which the influence of one stroke exceeds the previous. These will also correspond with salient points along the trajectory (extrema of curvature) and can simply be computed by finding the time occurrence at which two consecutive lognormals intersect. Similarly to the input key-point case, $\bm{\xi}(z_1)$ and $\bm{\xi}(z_m)$ respectively denote the first and last points of the generated trajectory.
\begin{figure}[h] 
\vspace{-3mm}
\centering
\includegraphics[width=0.5\textwidth]{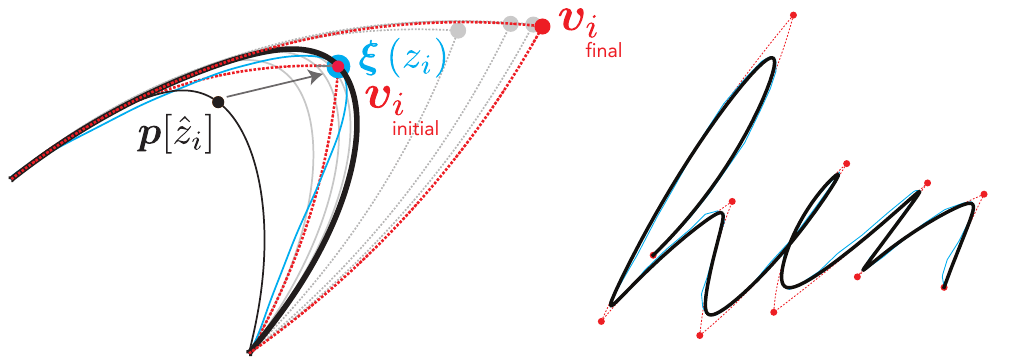} 
\caption{Final trajectory reconstruction step. Left, iterative adjustment of virtual target positions. Right, the final trajectory generated with the reconstructed dynamic parameters.  }
\label{fig:nudge}
\end{figure}
We then iteratively adjust the virtual target positions in order to move each generated key-point $\bm{\xi}(z_i)$ towards the corresponding input key-point $\bm{p}[\hat{z}_i]$ with:
	\begin{equation}
		\bm{v}_i \leftarrow \bm{v}_i + \bm{p}\left[\hat{z}_i\right] - \bm{\xi}\left(z_i\right),
	\end{equation}
The iteration continues until the Mean Square Error (MSE) of the distances between every pair $\bm{p}\left[\hat{z}_i\right]$ and $\bm{\xi}(z_i)$ is less than an experimentally set threshold or until a maximum number of iterations is reached (Fig. \ref{fig:nudge}). This method usually converges to a good reconstruction of the input within few iterations (experimentally $<5$). Interestingly, even though the dynamic information of the input is discarded, the reconstructed velocity profile is typically similar to the original in the number and outline of peaks (\refig{fig:param-recons}), which may be related to the extensively studied relationships existing between geometry and dynamics of movement trajectories~\cite{Flash2007}.

\subsection{Data augmentation}
\label{sect:augment}

\begin{figure}[h] 
\centering
\includegraphics[width=1.0\columnwidth]{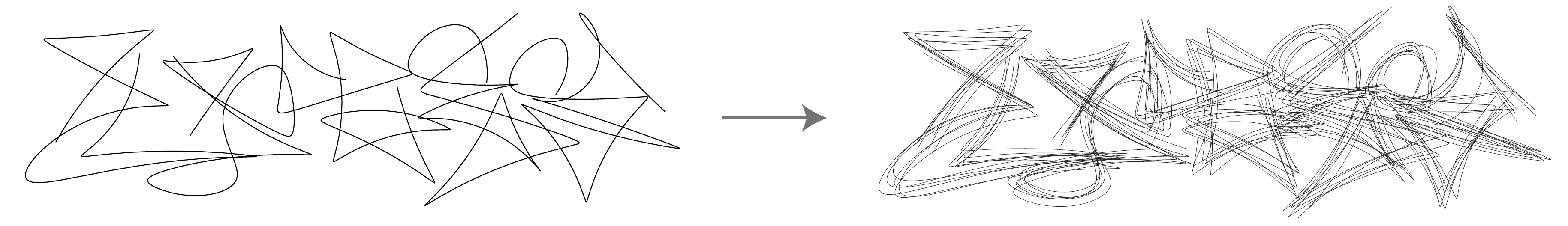} 
\caption{Data augmentation step.}
\vspace{-3 mm}
\label{fig:augment}
\end{figure}

We can exploit the $\Sigma \Lambda$ parameterisation to generate many variations over a single trajectory, which are visually consistent with the original, and with a variability that is similar to the one that would be seen in multiple instances of handwriting made by the same writer (\refig{fig:augment})
\cite{Fischer2014,berio2015cae}. Given a dataset of $n$ training samples, we randomly perturb the virtual target positions and dynamic parameters of each sample $n_p$ times, which results in a new augmented dataset of size $n + n \times n_p$ where legibility and trajectory smoothness is maintained across samples. This would not be possible on the original input dataset alone, as perturbations for each data-point would eventually result in a noisy unrecognisable trajectory.

\subsection{Dynamic Parameter Prediction (DPP)}
\label{sect:DPP}

Given a sequence of virtual targets, we would like to predict the corresponding dynamic parameters. We train a model to learn the probability distribution of the dynamic parameters $\{\Delta t_{0i}, \theta_i\}$ for the $i$th stroke, conditioned on the virtual targets and dynamic parameters leading up to that stroke:
\begin{equation}
\Pr(\Delta t_{0i}, \theta_i \given \vec{\Delta v}_{1:i}, u_{1:i}, \Delta t_{0(1:i-1)}, \theta_{1:i-1}) ,
\end{equation}
where $\vec{\Delta v}_i \in \reals^2$ denotes the relative position displacement (between the $i$th virtual target and the next), $u_i \in \{0, 1\}$ denotes the pen-up state, and the subscript notation $a:b$ denotes all indices from $a$ to $b$ inclusive. Note that the distribution is conditioned on the dynamic parameters of the \textit{previous stroke} $(\Delta t_{0i-1}, \theta_{i-1})$. Conceptually this represents a writer that knows their next virtual target, is aware of the dynamic history of their movement so far, and wants to know what dynamic parameters to use. 

We also considered an alternative model conditioned only on the virtual targets, independent of the previous dynamic parameters, i.e. $\Pr(\Delta t_{0i}, \theta_i \given \vec{\Delta v}_{1:i}, u_{1:i})$, which conceptually represents a writer unaware of the dynamic history of their movement. In our preliminary studies we found this model to not perform as well.

We implement our model using \textit{Recurrent Mixture Density Networks (RMDN)} with the LSTM architecture. An MDN \cite{Bishop1994} models and predicts the parameters of a \textit{Gaussian Mixture Model (GMM)}, i.e. a vector of means, covariances and mixture weights. An RMDN \cite{Schuster1999} is a combination of an RNN with an MDN, and outputs a \textit{unique set of GMM parameters at each timestep}, allowing the probability distribution to change with time as the input sequence develops.

\begin{figure}[h]
	\vspace{-2mm}
	\centering
	\includegraphics[width=0.5\textwidth]{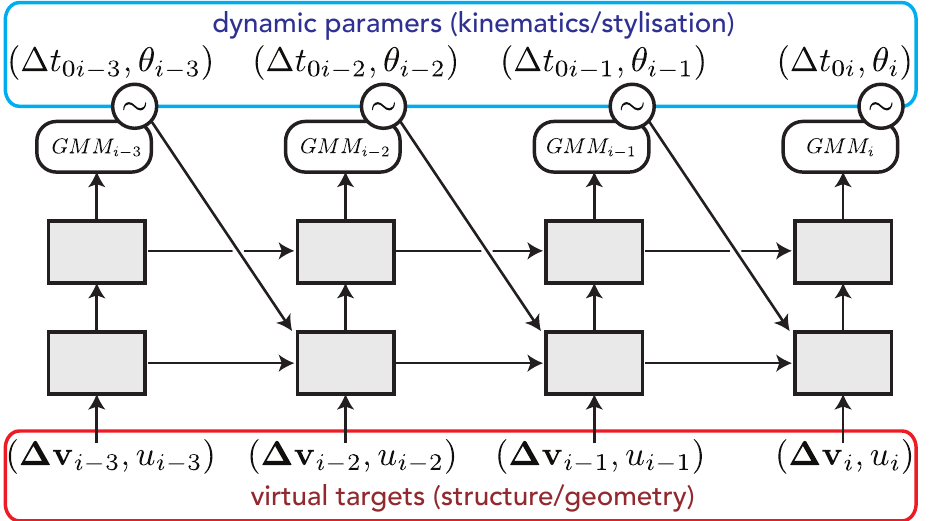} 
	\caption{Network architecture with two recurrent hidden layers. At each timestep $i$ the network outputs the parameters of a GMM which is sampled (denoted by $\sim$) and fed as input at the next timestep. }
	\label{fig:dpp_arch}
\end{figure}
At each timestep $i$, the input to the model is given by a vector $\vec{x}_i \in \reals^5$ corresponding to $[\vec{\Delta v}_i^\trsp, u_i, \Delta t_{0(i-1)}, \theta_{i-1}]$ where all elements (except for $u_i$) are normalised to zero mean and unit standard deviation. Our desired output is $\vec{z}_i \in \reals^2$ which consists of the normalised dynamic parameters for the $i$'th stroke $\{\Delta t_{0i}, \theta_i\}$ (\refig{fig:dpp_arch}).
Since $\vec{z}_i$ can be expressed as a \textit{bivariate} GMM, we can use a network architecture with output dimensions of $6 K$ where $K$ represents the number of Gaussians. This results in an output vector
$[\vec{\hat{\mu}}_i \in \reals^{2K},
\vec{\hat{\sigma}}_i \in \reals^{2K},
\vec{\hat{\rho}}_i \in \reals^{K} ,
\vec{\hat{\pi}}_i \in \reals^{K}]$ which we use to calculate the parameters of the GMM with (after Graves \cite{Graves2013}):
\begin{align}
\label{eq:rnnseq_rmdn_outputs}
\begin{split}
\vec{\mu}^k_i &= \vec{\hat{\mu}}_i^k : \text{means for $k$'th Gaussian, } \vec{\mu}^k_i \in \reals^2 ,\\
\vec{\sigma}^k_i &= \exp(\vec{\hat{\sigma}}^k_i) : \text{standard deviations for $k$'th Gaussian, } \vec{\sigma}^k_i \in \reals^2, \\
\rho^k_i &= \tanh(\hat{\rho}^k_i) : \text{correlations for $k$'th Gaussian, } \rho^k_i \in (-1, 1), \\
\pi^k_i &= \mathrm{softmax}(\hat{\pi}^k_i) : \text{mix weight for $k$'th Gaussian , } \sum_{k}^{K} \pi^k_i = 1.
\end{split}
\end{align}
We then formulate the probability of dynamic parameters $\vec{y}_i$ at timestep $i$, given input vector $\vec{x}_i$ as:

\begin{align}
\label{eq:rnnseq_rmdn_pdf}
P_i(\vec{y}_i \given \vec{x}_i) = \sum_{k}^{K} \pi^k_i \mathcal{N}(\vec{y}_{i} \given \vec{\mu}^k_i, \vec{\sigma}^k_i, \rho^k_i) .
\end{align}

\noindent We efficiently compute the bivariate Gaussian without requiring matrix inversions by exploiting its closed form solution \cite{Graves2013}. 
As our training objective, we use the \textit{Maximum Likelihood}, minimising the \textit{Negative Log Likelihood} (also known as \textit{Hamiltonian} or \textit{surprisal} \cite{Lin2016}). Given a training set of input-target pairs $(\vec{x}, \vec{\hat{y}})$, we define the loss for a single training example with:
\begin{equation}
J_s = 
- \sum_{i}^{L}
\ln P_i(\vec{\hat{y}} \given \vec{x}) ,
\label{eq:rnnseq_rnn_loss}
\end{equation}
where $L$ is the length of the sequence. The total loss is given by summing $J_s$ over all training examples. 

%
%
%
We use a form of Truncated Backpropagation Through Time (BPTT) \cite{Sutskever2013} whereby we segment long sequences into \textit{overlapping} segments of maximum length $L$. In this case long-term dependencies greater than length $L$ are lost, however with enough overlap the network can effectively learn a \textit{sliding window} of $L$ timesteps. We shuffle our training data and reset the internal state after each sequence. We empirically found an overlap factor of 50\% to perform well, though further studies are needed to confirm the sensitivity of this choice. We use \textit{dynamic unrolling} of the RNN, whereby the number of timesteps to unroll to is not set at compile time, in the architecture of the network, but unrolled dynamically while training, allowing variable length sequences. We train using the Adam optimizer \cite{kingma2014adam} with the recommended hyperparameters. To prevent exploding gradients we clip these using the L2 norm as described by Pascanu et al. \cite{Pascanu2013} and experimentally set the threshold.

We use LSTM networks \cite{Hochreiter1997}  with input, output and forget gates \cite{gers2000learning}, and dropout regularization \cite{Pham2014}. We have employed both a grid and a random search \cite{bergstra2012random} on various hyperparameters in the ranges: sequence length (5-128), number of hidden recurrent layers (1-3), dimensions per hidden layer (64-1024), number of Gaussians (5-20), dropout keep probability (50\%-95\%) and peepholes \{with, without\}. We experimentally settled on an architecture of 2 recurrent layers, each with size 400, 20 Gaussians, dropout keep probability of 90\% and no peepholes. 

\section{Experiments and Results}
\label{sect:results}
Given a sequence of virtual targets, the DPP model (\refsect{sect:DPP}) is used to produce different stylisations. The virtual targets can either be (i) directly defined by a user, (ii) reconstructed from an input path, or (iii) procedurally generated.
\begin{figure}[h!] 
	\centering
	\vspace{-2mm}
	\includegraphics[width=0.5\textwidth]{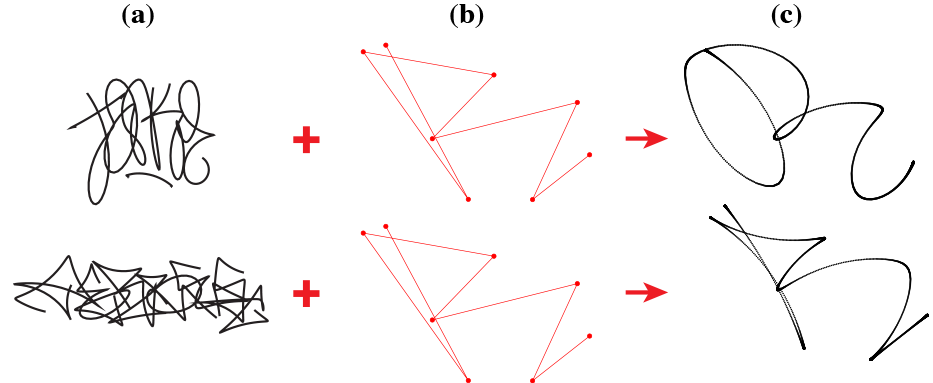} 
	\caption{Dynamic parameters generated over user specified virtual targets (in red), using \textit{two separate models}, each trained on a single example (with data augmentation x$8000$). Each row shows: \textbf{(a)} the training example, \textbf{(b)} the user provided virtual targets  and \textbf{(c)} the trajectory predicted by the corresponding model. Training examples: top row is from the GML database, bottom row was drawn using a tablet.}
	\label{fig:user_vt}
\end{figure}

\subsection{User defined virtual targets.} In this application of the system, a user can quickly sketch trajectories by interactively defining a sparse sequence of virtual targets. The system will then predict various smooth trajectories, the stylisation of which varies according to the dynamic parameters learned from a given training set. While the training procedure is performed offline, usually taking a few hours on a typical GPU, sampling the trained model runs at interactive rates. As a result, the user is able to view the results and interact in real time, for example by dragging the virtual targets around with a mouse, and seeing the final smooth trajectory update instantaneously with desired styles. 

\noindent\textbf{One-shot learning.}
The training set can contain one or more examples of target styles. As mentioned previously, we can use data augmentation to train a DPP model on as few as a \textit{single} initial training example, \textit{aka} one-shot learning, and the model is able to consistently predict dynamic parameters in that style (\refig{fig:user_vt}). 

\begin{figure}[h] 
	\centering
	\vspace{-2mm}
	\includegraphics[width=0.47\textwidth]{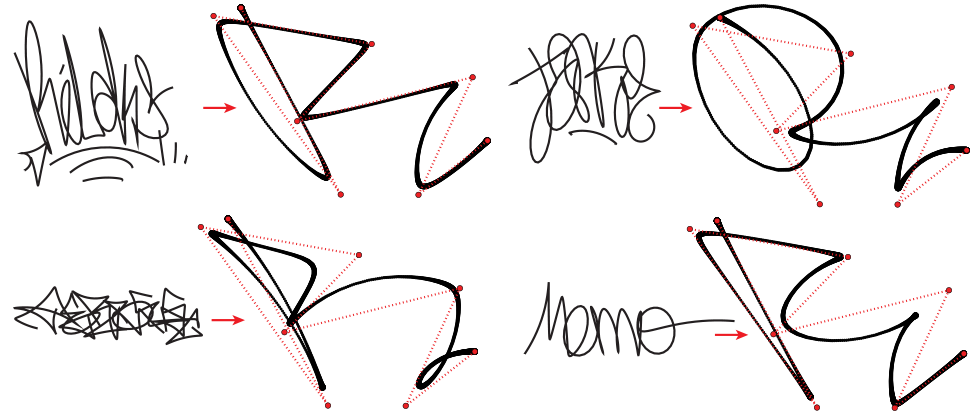} 
	\caption{Dynamic parameters generated over user specified virtual targets (in red) using \textit{a single model} trained on 4 examples (with data augmentation x$2000$). Each trajectory has been generated by priming the network with the corresponding example. Training examples: top row are from the GML database, bottom row were drawn using a tablet.}
	\label{fig:user_vt_multiple}
\end{figure}

\noindent\textbf{Priming}
We also train a DPP model on a data-set that contains multiple styles. In this case we can control the stylisation by \textit{priming} the model with the specific training example corresponding to the desired style (\refig{fig:user_vt_multiple}). Priming \cite{Graves2013} is achieved by first feeding the model a training example with the desired style, before feeding it the virtual targets which we wish to make a prediction on. We observe that a shorter sequence length (e.g. $15$) is sufficient for models trained on a single style example. However, for models trained on multiple styles, it is necessary to use a longer sequence length (e.g. $64$) to help it ``remember" the primed style across more virtual targets. 
 
\begin{figure}[h] 
 	\centering
 	\includegraphics[width=0.47\textwidth]{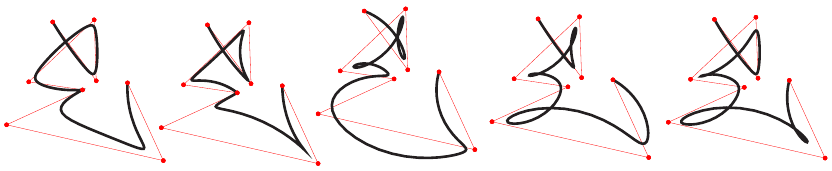} 
 	\caption{Variations generated by varying the random number generator seed prior to sampling. }
 	\label{fig:seed_variations}
 \end{figure}

 Because the model predicts parameters by stochastically sampling a learned distribution, different predictions and consequent variations of the generated trajectory can be achieved by choosing different seeds for the pseudo-random number generator (\refig{fig:seed_variations}).

\subsection{Trajectory stylisation.} 
We are also able to apply these methods to input paths previously drawn by a user or taken from an existing online dataset. Given such an input path, first we use our $\Sigma \Lambda$ parameter reconstruction method to extract a series of virtual targets and corresponding dynamic parameters that reconstruct the input path. We then \textit{discard} the reconstructed dynamic parameters and replace them with the ones predicted by the same DPP models we used in the previous examples. The result is an output that is structurally similar to the original path, but possesses the dynamic and geometric features of another.

\begin{figure}[ht] 
	\centering
	\includegraphics[width=0.5\textwidth]{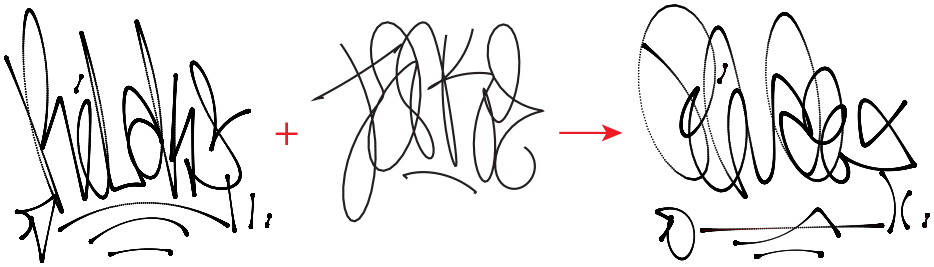} 
	\caption{Less satisfactory case for the stylisation of a complex tag. Left: user drawn trace to be stylised, Middle: training example, Right: Output of the model, applying the style of the training example to the user drawn trace.}
	\label{fig:style_transfer_fail}
\end{figure}

\begin{figure}[ht] 
	\centering
	\includegraphics[width=0.5\textwidth]{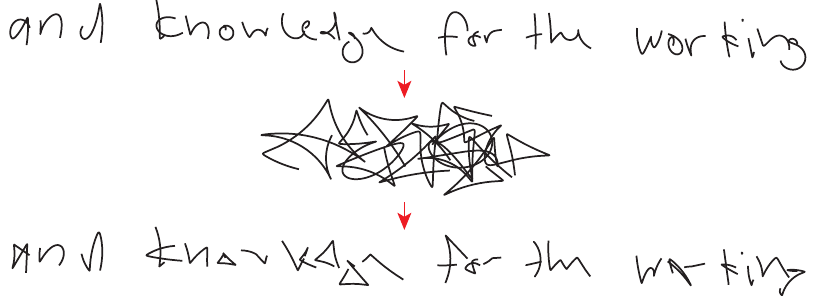} 
	\caption{Stylisation on longer sequences. Top: a sequence from the IAM dataset \citep{IamOnline2010}. Middle: single user-specificed sample used to train a DPP model. Bottom: final output.}
	\label{fig:style_transfer_iam}
\end{figure}

We test the method with a simple sequence of letters drawn by a user with a tablet and observe that, depending on the example used to prime the network, the method produces clearly different and readable stylisations of the input (\refig{fig:style_transfer}). On the other hand, the quality of the results strongly depends on the structural complexity of the input and on the perceptual similarity of the reconstructed virtual target sequence to the input path. In fact, with a more complex example, the original readability of the input may be lost and the quality of the resulting stylisation is not satisfactory, or in general difficult to evaluate qualitatively (\refig{fig:style_transfer_fail}). We also try this method on longer sequences of user-drawn traces such as those found in the IAM dataset \cite{IamOnline2010} (\refig{fig:style_transfer_iam}). 

\subsection{Fully generative traces}
Finally, we explore the application of our method in a generative pipeline in which a model learns and generates virtual target sequences, while the final trajectory and different stylisations are determined by the DPP model itself. For this task, we train a \textit{Virtual Target Prediction} (VTP) model, which builds from the handwriting prediction model originally developed by Graves \cite{Graves2013}. However, our VTP model implementation operates on \textit{sparse sequences of virtual targets} rather than dense raw point sequences. Due to this sparsity, it is then possible to augment the input in a coherent manner by perturbing the virtual targets positions with random offsets relative to the sequence extent. As a result, we can also train the model from a very small number of samples.

\begin{figure}[ht] 
	\centering
	\includegraphics[width=0.5\textwidth]{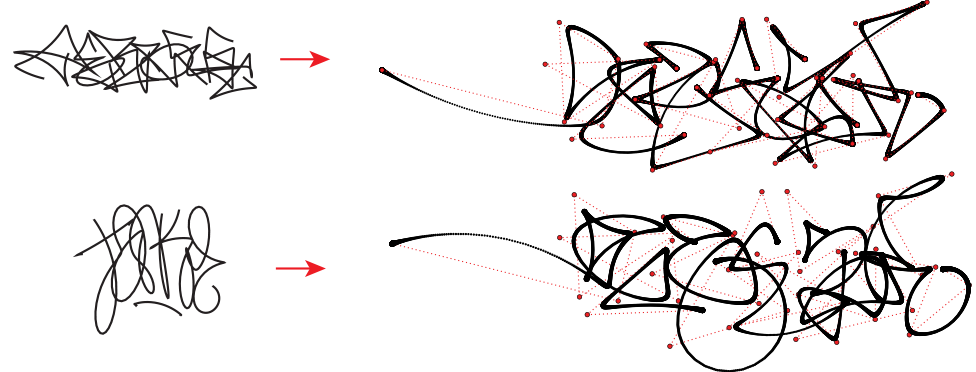} 
	\caption{Generating asemic graffiti. One network predicts sequences of virtual targets based on a small number of examples given by a user. Our DPP models are then used to predict the dynamics, resulting in different stylizations.}
	\label{fig:asemic}
\end{figure}

The process is as follows: training is performed on a first VTP model from virtual target sequences extracted from a small number of user made graffiti examples (e.g. between 1 and 4); we then use these models in combination to generate new virtual target sequences and patterns that are similar to the ones given in the training set. By priming the VTP model with a desired virtual target style example, and priming the DPP model with a desired dynamic-parameter style example (as in \refig{fig:user_vt_multiple}), it is then possible to control and mix the overall style of the generated trace (\refig{fig:asemic}). This process can also be directly used with large input datasets; e.g. we have trained VTP models on larger numbers of targets extracted from the IAM online handwriting dataset \cite{IamOnline2010} (as originally used by Graves \cite{Graves2013}).

\section{Conclusion}
\label{sect:conclusion}

We have demonstrated how an RNN-based architecture combined with a physiologically plausible model of human movement, the Sigma Lognormal ($\Sigma \Lambda$), can be used to implement a data driven path stylisation system. Our method functions similarly to existing path stylisation methods used in computer graphics applications 
(e.g. \cite{hertzmann2002curve,lang2015markov}).
However, we propose a movement based approach to stylisation, in which different styles are given by kinematic variations over a common structure of a hand drawn or written trace. We argue that using a physiologically plausible model of movement as a feature representation then provides a number of advantages with respect to polygonal (point sequence), or spline/interpolation based approaches. For example, the output of our method reflects a realistic and natural movement which can be used to (i) produce more expressive renderings of the trajectories by generating realistic stroke animations or (ii) drive the natural motions of an animated character or robotic drawing device. Furthermore, the variation of model parameters can be used (as demonstrated) to augment training data, but also to generate realistic variations in the generated outputs.

Our method currently relies on an accurate reconstruction of the input in the preprocessing step. Improvements should target especially parts that depend on user tuned parameters, such as the identification of salient points along the input trace (which requires a final peak detection pass), and measuring the sharpness of the input in correspondence with salient points. 

The reported work provides a solid basis for a number of different future research avenues. As an example, we hypothesize that a similar feature representation and architecture can be used to achieve handwriting synthesis results equivalent to the ones demonstrated by Graves \cite{Graves2013}, with the additional benefits of resolution independence and the possibility of training on a much reduced dataset size, even achieving satisfying results with one-shot learning situations.

\begin{acks}
The system takes as a starting point the original work developed by Graves \cite{Graves2013} and a public domain implementation by Ha \cite{otoro2015handwriting}. This work has been partly supported by UK's EPSRC Centre for Doctoral Training in Intelligent Games and Game Intelligence (IGGI; grant EP/L015846/1).
\end{acks}

	
\bibliographystyle{ACM-Reference-Format}
\bibliography{autograff-biblio.bib}


\begin{thebibliography}{00}


\ifx \showCODEN    \undefined \def \showCODEN     #1{\unskip}     \fi
\ifx \showDOI      \undefined \def \showDOI       #1{{\tt DOI:}\penalty0{#1}\ }
  \fi
\ifx \showISBNx    \undefined \def \showISBNx     #1{\unskip}     \fi
\ifx \showISBNxiii \undefined \def \showISBNxiii  #1{\unskip}     \fi
\ifx \showISSN     \undefined \def \showISSN      #1{\unskip}     \fi
\ifx \showLCCN     \undefined \def \showLCCN      #1{\unskip}     \fi
\ifx \shownote     \undefined \def \shownote      #1{#1}          \fi
\ifx \showarticletitle \undefined \def \showarticletitle #1{#1}   \fi
\ifx \showURL      \undefined \def \showURL       #1{#1}          \fi
\providecommand\bibfield[2]{#2}
\providecommand\bibinfo[2]{#2}
\providecommand\natexlab[1]{#1}
\providecommand\showeprint[2][]{arXiv:#2}

\bibitem[\protect\citeauthoryear{Bergstra and Bengio}{Bergstra and
  Bengio}{2012}]%
        {bergstra2012random}
\bibfield{author}{\bibinfo{person}{J. Bergstra} {and} \bibinfo{person}{Y.
  Bengio}.} \bibinfo{year}{2012}\natexlab{}.
\newblock \showarticletitle{{Random Search for Hyper-Parameter Optimization}}.
\newblock \bibinfo{journal}{{\em Journal of Machine Learning Research\/}}
  \bibinfo{volume}{13} (\bibinfo{year}{2012}), \bibinfo{pages}{281--305}.
\newblock


\bibitem[\protect\citeauthoryear{Berio, Calinon, and Fol~Leymarie}{Berio
  et~al\mbox{.}}{2016}]%
        {Berio16IROS}
\bibfield{author}{\bibinfo{person}{D. Berio}, \bibinfo{person}{S. Calinon},
  {and} \bibinfo{person}{F. Fol~Leymarie}.} \bibinfo{year}{2016}\natexlab{}.
\newblock \showarticletitle{Learning dynamic graffiti strokes with a compliant
  robot}. In \bibinfo{booktitle}{{\em Proc. {IEEE/RSJ} Intl Conf. on
  Intelligent Robots and Systems ({IROS})}}. \bibinfo{address}{Daejeon, Korea},
  \bibinfo{pages}{3981--6}.
\newblock
\newblock
\shownote{https://doi.org/10.1109/IROS.2016.7759586.}


\bibitem[\protect\citeauthoryear{Berio, Calinon, and Fol~Leymarie}{Berio
  et~al\mbox{.}}{2017a}]%
        {BerioMPCMOCO2017}
\bibfield{author}{\bibinfo{person}{D Berio}, \bibinfo{person}{S. Calinon},
  {and} \bibinfo{person}{F. Fol~Leymarie}.} \bibinfo{year}{2017}\natexlab{a}.
\newblock \showarticletitle{Dynamic Graffiti Stylisation with Stochastic
  Optimal Control}. In \bibinfo{booktitle}{{\em ACM Proceedings of the 4th
  International Conference on Movement and Computing}}.
  \bibinfo{address}{London, UK}.
\newblock


\bibitem[\protect\citeauthoryear{Berio, Calinon, and Fol~Leymarie}{Berio
  et~al\mbox{.}}{2017b}]%
        {BerioGI2017}
\bibfield{author}{\bibinfo{person}{D Berio}, \bibinfo{person}{S. Calinon},
  {and} \bibinfo{person}{F. Fol~Leymarie}.} \bibinfo{year}{2017}\natexlab{b}.
\newblock \showarticletitle{Generating Calligraphic Trajectories with Model
  Predictive Control}. In \bibinfo{booktitle}{{\em Proceedings of Graphics
  Interface}}. \bibinfo{publisher}{Canadian Human-Computer Communications
  Society}, \bibinfo{address}{Edmonton, Canada}.
\newblock


\bibitem[\protect\citeauthoryear{Berio and Fol~Leymarie}{Berio and
  Fol~Leymarie}{2015}]%
        {berio2015cae}
\bibfield{author}{\bibinfo{person}{D. Berio} {and} \bibinfo{person}{F.
  Fol~Leymarie}.} \bibinfo{year}{2015}\natexlab{}.
\newblock \showarticletitle{Computational Models for the Analysis and Synthesis
  of Graffiti Tag Strokes}. In \bibinfo{booktitle}{{\em Proceedings of the
  Workshop on Computational Aesthetics}}. \bibinfo{publisher}{Eurographics
  Association}, \bibinfo{pages}{35--47}.
\newblock


\bibitem[\protect\citeauthoryear{Bezine, Alimi, and Sherkat}{Bezine
  et~al\mbox{.}}{2004}]%
        {bezine2004beta}
\bibfield{author}{\bibinfo{person}{H. Bezine}, \bibinfo{person}{A.M. Alimi},
  {and} \bibinfo{person}{N. Sherkat}.} \bibinfo{year}{2004}\natexlab{}.
\newblock \showarticletitle{Generation and analysis of handwriting script with
  the {B}eta-elliptic model}.
\newblock \bibinfo{journal}{{\em Proc. of International Workshop on Frontiers
  in Handwriting Recognition (IWFHR)\/}} \bibinfo{volume}{8},
  \bibinfo{number}{2} (\bibinfo{year}{2004}), \bibinfo{pages}{515--20}.
\newblock


\bibitem[\protect\citeauthoryear{Bishop}{Bishop}{1994}]%
        {Bishop1994}
\bibfield{author}{\bibinfo{person}{C.M. Bishop}.}
  \bibinfo{year}{1994}\natexlab{}.
\newblock \showarticletitle{Mixture density networks}.
\newblock  (\bibinfo{year}{1994}).
\newblock
\newblock
\shownote{Technical Report NCRG/4288, Neural Computing Research Group, Aston
  University, U.K.}


\bibitem[\protect\citeauthoryear{Crnkovic{-}Friis and
  Crnkovic{-}Friis}{Crnkovic{-}Friis and Crnkovic{-}Friis}{2016}]%
        {friis2016}
\bibfield{author}{\bibinfo{person}{L. Crnkovic{-}Friis} {and}
  \bibinfo{person}{L. Crnkovic{-}Friis}.} \bibinfo{year}{2016}\natexlab{}.
\newblock \showarticletitle{Generative Choreography using Deep Learning}.
\newblock \bibinfo{journal}{{\em arXiv preprint arXiv:1605.06921\/}}
  (\bibinfo{year}{2016}).
\newblock


\bibitem[\protect\citeauthoryear{D. and J.}{D. and J.}{2014}]%
        {kingma2014adam}
\bibfield{author}{\bibinfo{person}{Kingma D.} {and} \bibinfo{person}{Ba J.}}
  \bibinfo{year}{2014}\natexlab{}.
\newblock \showarticletitle{{Adam: A method for stochastic optimization}}.
\newblock \bibinfo{journal}{{\em arXiv preprint arXiv:1412.6980\/}}
  (\bibinfo{year}{2014}), \bibinfo{pages}{1--13}.
\newblock


\bibitem[\protect\citeauthoryear{De~Winter and Wagemans}{De~Winter and
  Wagemans}{2008}]%
        {dewinter2008perceptual}
\bibfield{author}{\bibinfo{person}{J. De~Winter} {and} \bibinfo{person}{J.
  Wagemans}.} \bibinfo{year}{2008}\natexlab{}.
\newblock \showarticletitle{Perceptual saliency of points along the contour of
  everyday objects: A large-scale study}.
\newblock \bibinfo{journal}{{\em Perception \& Psychophysics\/}}
  \bibinfo{volume}{70}, \bibinfo{number}{1} (\bibinfo{year}{2008}),
  \bibinfo{pages}{50--64}.
\newblock


\bibitem[\protect\citeauthoryear{Dempster, Laird, and Rubin}{Dempster
  et~al\mbox{.}}{1977}]%
        {Dempster77}
\bibfield{author}{\bibinfo{person}{A. Dempster}, \bibinfo{person}{N. Laird},
  {and} \bibinfo{person}{D. Rubin}.} \bibinfo{year}{1977}\natexlab{}.
\newblock \showarticletitle{Maximum Likelihood from Incomplete Data via the
  {EM} Algorithm}.
\newblock \bibinfo{journal}{{\em J. Royal Statistical Society B\/}}
  \bibinfo{volume}{39}, \bibinfo{number}{1} (\bibinfo{year}{1977}),
  \bibinfo{pages}{1--38}.
\newblock


\bibitem[\protect\citeauthoryear{Feldman and Singh}{Feldman and Singh}{2005}]%
        {feldman2005information}
\bibfield{author}{\bibinfo{person}{J. Feldman} {and} \bibinfo{person}{M.
  Singh}.} \bibinfo{year}{2005}\natexlab{}.
\newblock \showarticletitle{Information along contours and object boundaries.}
\newblock \bibinfo{journal}{{\em Psychological review\/}}
  \bibinfo{volume}{112}, \bibinfo{number}{1} (\bibinfo{year}{2005}),
  \bibinfo{pages}{243}.
\newblock


\bibitem[\protect\citeauthoryear{Fischer, Plamondon, O'Reilly, and
  Savaria}{Fischer et~al\mbox{.}}{2014}]%
        {Fischer2014}
\bibfield{author}{\bibinfo{person}{A. Fischer}, \bibinfo{person}{R. Plamondon},
  \bibinfo{person}{C. O'Reilly}, {and} \bibinfo{person}{Y. Savaria}.}
  \bibinfo{year}{2014}\natexlab{}.
\newblock \showarticletitle{Neuromuscular Representation and Synthetic
  Generation of Handwritten Whiteboard Notes}. In \bibinfo{booktitle}{{\em
  Proc. of Frontiers in Handwriting Recognition (ICFHR)}}.
  \bibinfo{pages}{222--7}.
\newblock


\bibitem[\protect\citeauthoryear{Flash and Handzel}{Flash and Handzel}{2007}]%
        {Flash2007}
\bibfield{author}{\bibinfo{person}{T. Flash} {and} \bibinfo{person}{A.A.
  Handzel}.} \bibinfo{year}{2007}\natexlab{}.
\newblock \showarticletitle{Affine differential geometry analysis of human arm
  movements}.
\newblock \bibinfo{journal}{{\em Biological cybernetics\/}}
  \bibinfo{volume}{96}, \bibinfo{number}{6} (\bibinfo{year}{2007}),
  \bibinfo{pages}{577--601}.
\newblock


\bibitem[\protect\citeauthoryear{Flash and Hochner}{Flash and Hochner}{2005}]%
        {Flash2005}
\bibfield{author}{\bibinfo{person}{T. Flash} {and} \bibinfo{person}{B.
  Hochner}.} \bibinfo{year}{2005}\natexlab{}.
\newblock \showarticletitle{Motor primitives in vertebrates and invertebrates}.
\newblock \bibinfo{journal}{{\em Current opinion in neurobiology\/}}
  \bibinfo{volume}{15}, \bibinfo{number}{6} (\bibinfo{year}{2005}),
  \bibinfo{pages}{660--6}.
\newblock


\bibitem[\protect\citeauthoryear{Gers, Schmidhuber, and Cummins}{Gers
  et~al\mbox{.}}{2000}]%
        {gers2000learning}
\bibfield{author}{\bibinfo{person}{F.A. Gers}, \bibinfo{person}{J.
  Schmidhuber}, {and} \bibinfo{person}{F. Cummins}.}
  \bibinfo{year}{2000}\natexlab{}.
\newblock \showarticletitle{{Learning to forget: Continual prediction with
  LSTM}}.
\newblock \bibinfo{journal}{{\em Neural Computation\/}} \bibinfo{volume}{12},
  \bibinfo{number}{10} (\bibinfo{year}{2000}), \bibinfo{pages}{2451--2471}.
\newblock


\bibitem[\protect\citeauthoryear{Graves}{Graves}{2013}]%
        {Graves2013}
\bibfield{author}{\bibinfo{person}{A. Graves}.}
  \bibinfo{year}{2013}\natexlab{}.
\newblock \showarticletitle{Generating sequences with recurrent neural
  networks}.
\newblock \bibinfo{journal}{{\em arXiv preprint arXiv:1308.0850\/}}
  (\bibinfo{year}{2013}).
\newblock


\bibitem[\protect\citeauthoryear{Gregor, Danihelka, Graves, and
  Wierstra}{Gregor et~al\mbox{.}}{2015}]%
        {Gregor2015}
\bibfield{author}{\bibinfo{person}{K. Gregor}, \bibinfo{person}{I. Danihelka},
  \bibinfo{person}{A. Graves}, {and} \bibinfo{person}{D. Wierstra}.}
  \bibinfo{year}{2015}\natexlab{}.
\newblock \showarticletitle{{DRAW: A Recurrent Neural Network For Image
  Generation}}.
\newblock \bibinfo{journal}{{\em arXiv preprint arXiv:1502.04623\/}}
  (\bibinfo{year}{2015}).
\newblock


\bibitem[\protect\citeauthoryear{Ha}{Ha}{2015}]%
        {otoro2015handwriting}
\bibfield{author}{\bibinfo{person}{D. Ha}.} \bibinfo{year}{2015}\natexlab{}.
\newblock \bibinfo{title}{{Generative Handwriting Demo using TensorFlow}}.
\newblock   (\bibinfo{year}{2015}).
\newblock
\showURL{%
\url{https://github.com/hardmaru/write-rnn-tensorflow}}


\bibitem[\protect\citeauthoryear{Ha, Dai, and Le}{Ha et~al\mbox{.}}{2016}]%
        {ha2016hypernetworks}
\bibfield{author}{\bibinfo{person}{D. Ha}, \bibinfo{person}{A. Dai}, {and}
  \bibinfo{person}{Q.V. Le}.} \bibinfo{year}{2016}\natexlab{}.
\newblock \showarticletitle{HyperNetworks}.
\newblock \bibinfo{journal}{{\em arXiv preprint arXiv:1609.09106\/}}
  (\bibinfo{year}{2016}).
\newblock


\bibitem[\protect\citeauthoryear{Hertzmann, Oliver, Curless, and
  Seitz}{Hertzmann et~al\mbox{.}}{2002}]%
        {hertzmann2002curve}
\bibfield{author}{\bibinfo{person}{A. Hertzmann}, \bibinfo{person}{N. Oliver},
  \bibinfo{person}{B. Curless}, {and} \bibinfo{person}{S.M. Seitz}.}
  \bibinfo{year}{2002}\natexlab{}.
\newblock \showarticletitle{Curve Analogies}. In \bibinfo{booktitle}{{\em Proc.
  13th EuroGraphics Workshop on Rendering (EGRW)}}. \bibinfo{address}{Pisa,
  Italy}, \bibinfo{pages}{233--46}.
\newblock


\bibitem[\protect\citeauthoryear{Hochreiter and Schmidhuber}{Hochreiter and
  Schmidhuber}{1997}]%
        {Hochreiter1997}
\bibfield{author}{\bibinfo{person}{S. Hochreiter} {and} \bibinfo{person}{J.
  Schmidhuber}.} \bibinfo{year}{1997}\natexlab{}.
\newblock \showarticletitle{Long Short-Term Memory}.
\newblock \bibinfo{journal}{{\em Neural Computation\/}} \bibinfo{volume}{9},
  \bibinfo{number}{8} (\bibinfo{year}{1997}), \bibinfo{pages}{1735--80}.
\newblock


\bibitem[\protect\citeauthoryear{Inderm{\"u}hle, Liwicki, and
  Bunke}{Inderm{\"u}hle et~al\mbox{.}}{2010}]%
        {IamOnline2010}
\bibfield{author}{\bibinfo{person}{E. Inderm{\"u}hle}, \bibinfo{person}{M.
  Liwicki}, {and} \bibinfo{person}{H. Bunke}.} \bibinfo{year}{2010}\natexlab{}.
\newblock \showarticletitle{{IAMonDo}-database: An Online Handwritten Document
  Database with Non-uniform Contents}. In \bibinfo{booktitle}{{\em Proc. 9th
  IAPR Int'l Workshop on Document Analysis Systems (DAS)}}.
  \bibinfo{pages}{97--104}.
\newblock


\bibitem[\protect\citeauthoryear{Kao, Hoosain, and Van~Galen}{Kao
  et~al\mbox{.}}{1986}]%
        {kao1986graphonomics}
\bibfield{author}{\bibinfo{person}{H.S.R. Kao}, \bibinfo{person}{R. Hoosain},
  {and} \bibinfo{person}{G.P. Van~Galen}.} \bibinfo{year}{1986}\natexlab{}.
\newblock \bibinfo{booktitle}{{\em Graphonomics: Contemporary research in
  handwriting}}.
\newblock \bibinfo{publisher}{Elsevier}.
\newblock


\bibitem[\protect\citeauthoryear{Lab}{Lab}{2009}]%
        {GraffitiAnalysis}
\bibfield{author}{\bibinfo{person}{Graffiti~Research Lab}.}
  \bibinfo{year}{2009}\natexlab{}.
\newblock \bibinfo{title}{Graffiti Analysis Database (000000book.com)}.
\newblock \bibinfo{howpublished}{Web}.   (\bibinfo{year}{2009}).
\newblock
\showURL{%
\url{http://000000book.com}}


\bibitem[\protect\citeauthoryear{Lang and Alexa}{Lang and Alexa}{2015}]%
        {lang2015markov}
\bibfield{author}{\bibinfo{person}{K. Lang} {and} \bibinfo{person}{M. Alexa}.}
  \bibinfo{year}{2015}\natexlab{}.
\newblock \showarticletitle{The {M}arkov pen: Online synthesis of free-hand
  drawing styles}. In \bibinfo{booktitle}{{\em Proc. of the workshop on
  Non-Photorealistic Animation and Rendering (NPAR)}}. Eurographics
  Association, \bibinfo{pages}{203--15}.
\newblock


\bibitem[\protect\citeauthoryear{Lin and Tegmark}{Lin and Tegmark}{2016}]%
        {Lin2016}
\bibfield{author}{\bibinfo{person}{H.W. Lin} {and} \bibinfo{person}{M.
  Tegmark}.} \bibinfo{year}{2016}\natexlab{}.
\newblock \showarticletitle{{Why does deep and cheap learning work so well?}}
\newblock \bibinfo{journal}{{\em arXiv preprint arXiv:1608.08225\/}}
  (\bibinfo{year}{2016}).
\newblock


\bibitem[\protect\citeauthoryear{Ltaief, Bezine, and Alimi}{Ltaief
  et~al\mbox{.}}{2012}]%
        {Ltaief2012}
\bibfield{author}{\bibinfo{person}{M. Ltaief}, \bibinfo{person}{H. Bezine},
  {and} \bibinfo{person}{A.M. Alimi}.} \bibinfo{year}{2012}\natexlab{}.
\newblock \showarticletitle{A neuro-{B}eta-elliptic model for handwriting
  generation movements}. In \bibinfo{booktitle}{{\em Proc. of International
  Conference on Frontiers in Handwriting Recognition (ICFHR)}}.
  \bibinfo{pages}{803--8}.
\newblock


\bibitem[\protect\citeauthoryear{Nagasaki}{Nagasaki}{1989}]%
        {Nagasaki1989}
\bibfield{author}{\bibinfo{person}{H. Nagasaki}.}
  \bibinfo{year}{1989}\natexlab{}.
\newblock \showarticletitle{Asymmetric velocity and acceleration profiles of
  human arm movements}.
\newblock \bibinfo{journal}{{\em Experimental Brain Research\/}}
  \bibinfo{volume}{74}, \bibinfo{number}{2} (\bibinfo{year}{1989}),
  \bibinfo{pages}{319--26}.
\newblock


\bibitem[\protect\citeauthoryear{Nair and Hinton}{Nair and Hinton}{2005}]%
        {Nair2005}
\bibfield{author}{\bibinfo{person}{V. Nair} {and} \bibinfo{person}{G. Hinton}.}
  \bibinfo{year}{2005}\natexlab{}.
\newblock \showarticletitle{Inferring motor programs from images of handwritten
  digits}. In \bibinfo{booktitle}{{\em Advances in neural information
  processing systems}}. \bibinfo{pages}{515--22}.
\newblock


\bibitem[\protect\citeauthoryear{O'Reilly and Plamondon}{O'Reilly and
  Plamondon}{2008}]%
        {OReilly2008}
\bibfield{author}{\bibinfo{person}{C. O'Reilly} {and} \bibinfo{person}{R.
  Plamondon}.} \bibinfo{year}{2008}\natexlab{}.
\newblock \showarticletitle{Automatic extraction of {S}igma-{L}ognormal
  parameters on signatures}. In \bibinfo{booktitle}{{\em Proc. of 11th
  International Conference on Frontier in Handwriting Recognition (ICFHR)}}.
\newblock


\bibitem[\protect\citeauthoryear{Pascanu, Mikolov, and Bengio}{Pascanu
  et~al\mbox{.}}{2013}]%
        {Pascanu2013}
\bibfield{author}{\bibinfo{person}{R. Pascanu}, \bibinfo{person}{T. Mikolov},
  {and} \bibinfo{person}{Y. Bengio}.} \bibinfo{year}{2013}\natexlab{}.
\newblock \showarticletitle{On the difficulty of training {R}ecurrent {N}eural
  {N}etworks}. In \bibinfo{booktitle}{{\em Proc. of ICML}},
  Vol.~\bibinfo{volume}{28}. \bibinfo{pages}{1310--8}.
\newblock


\bibitem[\protect\citeauthoryear{Pham, Bluche, Kermorvant, and Louradour}{Pham
  et~al\mbox{.}}{2014}]%
        {Pham2014}
\bibfield{author}{\bibinfo{person}{V. Pham}, \bibinfo{person}{T. Bluche},
  \bibinfo{person}{C. Kermorvant}, {and} \bibinfo{person}{J. Louradour}.}
  \bibinfo{year}{2014}\natexlab{}.
\newblock \showarticletitle{{Dropout Improves Recurrent Neural Networks for
  Handwriting Recognition}}. In \bibinfo{booktitle}{{\em Proc. of ICFHR}}.
  \bibinfo{publisher}{IEEE}, \bibinfo{pages}{285--90}.
\newblock


\bibitem[\protect\citeauthoryear{Pignocchi}{Pignocchi}{2010}]%
        {pignocchi2010}
\bibfield{author}{\bibinfo{person}{A. Pignocchi}.}
  \bibinfo{year}{2010}\natexlab{}.
\newblock \showarticletitle{{How the Intentions of the Draftsman Shape
  Perception of a Drawing}}.
\newblock \bibinfo{journal}{{\em Consciousness and Cognition\/}}
  \bibinfo{volume}{19}, \bibinfo{number}{4} (\bibinfo{year}{2010}),
  \bibinfo{pages}{887--98}.
\newblock


\bibitem[\protect\citeauthoryear{Plamondon}{Plamondon}{1995}]%
        {plamondon1995kinematic}
\bibfield{author}{\bibinfo{person}{R. Plamondon}.}
  \bibinfo{year}{1995}\natexlab{}.
\newblock \showarticletitle{{A Kinematic Theory of Rapid Human Movements. Part
  I . Movement Representation and Generation}}.
\newblock \bibinfo{journal}{{\em Biological cybernetics\/}}
  \bibinfo{volume}{72}, \bibinfo{number}{4} (\bibinfo{year}{1995}),
  \bibinfo{pages}{295--307}.
\newblock


\bibitem[\protect\citeauthoryear{Plamondon et~al\mbox{.}}{Plamondon
  et~al\mbox{.}}{1993}]%
        {Plamondon1993}
\bibfield{author}{\bibinfo{person}{R. Plamondon} {and}
  \bibinfo{person}{others}.} \bibinfo{year}{1993}\natexlab{}.
\newblock \showarticletitle{Modelling velocity profiles of rapid movements: A
  comparative study}.
\newblock \bibinfo{journal}{{\em Biological cybernetics\/}}
  \bibinfo{volume}{69}, \bibinfo{number}{2} (\bibinfo{year}{1993}),
  \bibinfo{pages}{119--28}.
\newblock


\bibitem[\protect\citeauthoryear{Plamondon et~al\mbox{.}}{Plamondon
  et~al\mbox{.}}{2014}]%
        {Plamondon2014}
\bibfield{author}{\bibinfo{person}{R. Plamondon} {and}
  \bibinfo{person}{others}.} \bibinfo{year}{2014}\natexlab{}.
\newblock \showarticletitle{Recent developments in the study of rapid human
  movements with the kinematic theory}.
\newblock \bibinfo{journal}{{\em Pattern Recognition Letters\/}}
  \bibinfo{volume}{35} (\bibinfo{year}{2014}), \bibinfo{pages}{225--35}.
\newblock


\bibitem[\protect\citeauthoryear{Plamondon and Privitera}{Plamondon and
  Privitera}{1996}]%
        {Plamondon1996}
\bibfield{author}{\bibinfo{person}{R. Plamondon} {and} \bibinfo{person}{C.
  Privitera}.} \bibinfo{year}{1996}\natexlab{}.
\newblock \showarticletitle{A neural model for generating and learning a rapid
  movement sequence}.
\newblock \bibinfo{journal}{{\em Biological cybernetics\/}}
  \bibinfo{volume}{74}, \bibinfo{number}{2} (\bibinfo{year}{1996}),
  \bibinfo{pages}{117--30}.
\newblock


\bibitem[\protect\citeauthoryear{Rohrer and Hogan}{Rohrer and Hogan}{2006}]%
        {Rohrer2006}
\bibfield{author}{\bibinfo{person}{B. Rohrer} {and} \bibinfo{person}{N.
  Hogan}.} \bibinfo{year}{2006}\natexlab{}.
\newblock \showarticletitle{Avoiding spurious submovement decompositions {II}}.
\newblock \bibinfo{journal}{{\em Biological cybernetics\/}}
  \bibinfo{volume}{94}, \bibinfo{number}{5} (\bibinfo{year}{2006}),
  \bibinfo{pages}{409--14}.
\newblock


\bibitem[\protect\citeauthoryear{Rosenbaum et~al\mbox{.}}{Rosenbaum
  et~al\mbox{.}}{1995}]%
        {Rosenbaum1995}
\bibfield{author}{\bibinfo{person}{D.A. Rosenbaum} {and}
  \bibinfo{person}{others}.} \bibinfo{year}{1995}\natexlab{}.
\newblock \showarticletitle{Planning reaches by evaluating stored postures}.
\newblock \bibinfo{journal}{{\em Psychological Review\/}}
  \bibinfo{volume}{102}, \bibinfo{number}{1} (\bibinfo{year}{1995}),
  \bibinfo{pages}{28--67}.
\newblock


\bibitem[\protect\citeauthoryear{Schuster}{Schuster}{1999}]%
        {Schuster1999}
\bibfield{author}{\bibinfo{person}{M. Schuster}.}
  \bibinfo{year}{1999}\natexlab{}.
\newblock \showarticletitle{{Better Generative Models for Sequential Data
  Problems: Bidirectional Recurrent Mixture Density Networks}}. In
  \bibinfo{booktitle}{{\em Proc. of NIPS}}. \bibinfo{pages}{589--95}.
\newblock


\bibitem[\protect\citeauthoryear{Sutskever}{Sutskever}{2013}]%
        {Sutskever2013}
\bibfield{author}{\bibinfo{person}{I. Sutskever}.}
  \bibinfo{year}{2013}\natexlab{}.
\newblock {\em \bibinfo{title}{{Training Recurrent Neural Networks}}}.
\newblock \bibinfo{thesistype}{Ph.D. Dissertation}. \bibinfo{school}{University
  of Toronto}.
\newblock


\bibitem[\protect\citeauthoryear{Umilta et~al\mbox{.}}{Umilta
  et~al\mbox{.}}{2012}]%
        {Umilta2012}
\bibfield{author}{\bibinfo{person}{M. Umilta} {and} \bibinfo{person}{others}.}
  \bibinfo{year}{2012}\natexlab{}.
\newblock \showarticletitle{Abstract Art and Cortical Motor Activation: An
  {EEG} Study}.
\newblock \bibinfo{journal}{{\em Frontiers in Human Neuroscience\/}}
  \bibinfo{volume}{6}, \bibinfo{number}{311} (\bibinfo{year}{2012}),
  \bibinfo{pages}{1--9}.
\newblock


\bibitem[\protect\citeauthoryear{Varga, Kilchhofer, and Bunke}{Varga
  et~al\mbox{.}}{2005}]%
        {varga2005template}
\bibfield{author}{\bibinfo{person}{T. Varga}, \bibinfo{person}{D. Kilchhofer},
  {and} \bibinfo{person}{H. Bunke}.} \bibinfo{year}{2005}\natexlab{}.
\newblock \showarticletitle{{Template-based Synthetic Handwriting Generation
  for the Training of Recognition Systems}}. In \bibinfo{booktitle}{{\em Proc.
  of 12th Conf. of the International Graphonomics Society (IGS)}}.
  \bibinfo{pages}{206--11}.
\newblock


\bibitem[\protect\citeauthoryear{Zhang et~al\mbox{.}}{Zhang
  et~al\mbox{.}}{2016}]%
        {Zhang2016}
\bibfield{author}{\bibinfo{person}{X.-Y. Zhang} {and}
  \bibinfo{person}{others}.} \bibinfo{year}{2016}\natexlab{}.
\newblock \showarticletitle{{Drawing and Recognizing Chinese Characters with
  Recurrent Neural Network}}.
\newblock \bibinfo{journal}{{\em arXiv preprint arXiv:1606.06539\/}}
  (\bibinfo{year}{2016}).
\newblock


\end{thebibliography}

\end{document}